\documentclass[letterpaper, 10 pt, conference]{ieeeconf}  

\IEEEoverridecommandlockouts                              

\usepackage{hyperref}
\usepackage{acro}
\acsetup{
	single-style = long,
	single = true
}
\usepackage{algorithm}
\usepackage{xcolor}
\usepackage{siunitx}
\usepackage{nicefrac}
\usepackage{tikz}
\usepackage[utf8]{inputenc}
\usepackage{pgfplots}
\DeclareUnicodeCharacter{2212}{−}
\usepgfplotslibrary{groupplots,dateplot}
\usetikzlibrary{patterns,shapes.arrows}
\pgfplotsset{compat=newest}
\usepackage{cite}
\usepackage{amsmath,amssymb,amsfonts}
\usepackage[noend]{algorithmic}
\usepackage{graphicx}
\usepackage{textcomp}
\usepackage{xcolor}
\usepackage{bm}

\usepackage{array}
\usepackage{diagbox}
\usepackage{multirow}
\makeatletter
\newcommand{\thickhline}{%
	\noalign {\ifnum 0=`}\fi \hrule height 1pt
	\futurelet \reserved@a \@xhline
}
\newcolumntype{"}{@{\hskip\tabcolsep\vrule width 1pt\hskip\tabcolsep}}
\makeatother
\usepackage{subcaption}
\DeclareAcronym{LQR}{
	short=LQR,
	long=linear-quadratic regulator
}

\DeclareAcronym{iLQR}{
	short=iLQR,
	long=iterative linear-quadratic regulator
}

\DeclareAcronym{iLQG}{
	short=iLQG,
	long=iterative linear-quadratic Gaussian
}

\DeclareAcronym{LQ}{
	short=LQ,
	long=linear-quadratic
}

\DeclareAcronym{CFR}{
	short=CFR,
	long=counterfactual regret
}

\DeclareAcronym{iLQGame}{
	short=iLQGames,
	long=iterative linear-quadratic games
}

\DeclareAcronym{OCP}{
	short=OCP,
	long=optimal control problem
}

\DeclareAcronym{POMDP}{
	short=POMDP,
	long=partially observable Markov desicion processes
}

\DeclareAcronym{IBR}{
	short=IBR,
	long=iterative best response
}

\DeclareAcronym{DDP}{
	short=DDP,
	long=differential dynamic programming
}

\DeclareAcronym{SQP}{
	short=SQP,
	long=sequential quadratic programming
}

\DeclareAcronym{KKT}{
	short=KKT,
	long=Karush–Kuhn–Tucker,
}
\DeclareAcronym{MPC}{
	short=MPC,
	long=model predictive control,
}
\usepackage{blindtext}  

\title{\LARGE \bf
Open-Loop and Feedback Nash Trajectories for Competitive Racing with iLQGames*
}

\author{Matthias Rowold$^{1}$, Alexander Langmann$^{2}$, Boris Lohmann$^{1}$, and Johannes Betz$^{2}$
\thanks{*This work was not supported by any organization}
\thanks{$^{1}$Matthias Rowold and Boris Lohmann are with the Chair of Automatic Control, TUM School of Engineering and Design, Technical University of Munich, Germany {\tt\small \{matthias.rowold, lohmann\}@tum.de}}
\thanks{$^{2}$Alexander Langmann and Johannes Betz are with the Professorship of Autonomous Vehicle Systems, TUM School of Engineering and Design, Technical University of Munich, Germany {\tt\small \{alexander.langmann, johannes.betz\}@tum.de}}
}

\begin{document}

\maketitle
\thispagestyle{empty}
\pagestyle{empty}

\begin{abstract}
Interaction-aware trajectory planning is crucial for closing the gap between autonomous racing cars and human racing drivers. Prior work has applied game theory as it provides equilibrium concepts for non-cooperative dynamic problems. With this contribution, we formulate racing as a dynamic game and employ a variant of iLQR---called iLQGames---to solve the game. iLQGames finds trajectories for all players that satisfy the equilibrium conditions for a linear-quadratic approximation of the game and has been previously applied in traffic scenarios. We analyze the algorithm's applicability for trajectory planning in racing scenarios and evaluate it based on interaction awareness, competitiveness, and safety. With the ability of iLQGames to solve for open-loop and feedback Nash equilibria, we compare the behavioral outcomes of the two equilibrium concepts in simple scenarios on a straight track section.
\end{abstract}

\section{Introduction}
\label{sec:introduction}
\begin{figure}[t]
	\small
	\centering
	\def\axiswidth{8.5cm}
	\input{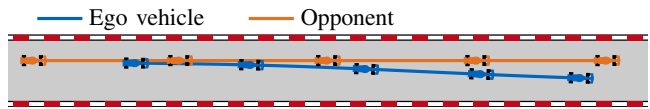}
	\caption{Example of yielding behavior by sequential planning approaches in racing scenarios.}
	\label{fig:yielding_behavior}
	\vspace{-0.5cm}
\end{figure}

Despite the progress of algorithms for autonomous racing cars in recent years, they cannot compete with human drivers in head-to-head races or even more complex multi-vehicle scenarios. Besides limitations imposed by sensor ranges and control performance, trajectory planning plays a crucial part in this discrepancy. Established planning approaches as in \cite{Rowold.2022, Raji.2022, Jank.2023}, which were applied in head-to-head races at the Indy Autonomous Challenge, do not achieve the competitive and strategic maneuvers characterizing racing between human drivers. They first predict the trajectories of opponents and then react with a collision-free trajectory. The prediction is assumed to be definitive, so this \emph{sequential} procedure fails to capture the reciprocal nature of the planning problem. This means that it neglects that the opponents will react to the executed motion of the ego vehicle and that the prediction itself depends on the planned trajectory. In racing, this neglect negatively impacts even in supposedly simple scenarios. For instance, the opponent in Fig.~\ref{fig:yielding_behavior} approaches the ego vehicle with a higher speed. Using a sequential planning approach, the ego vehicle predicts the opponent to maintain its speed on a straight path and consequently makes way to avoid a collision. A competitive human driver would anticipate an overtaking maneuver and stay on the left side or even try blocking the opponent.

Like in the example, sequential approaches often lead to overly cautious trajectories, and they can even result in the ego vehicle becoming immobilized in scenarios with rapidly increasing prediction uncertainties. This phenomenon is commonly referred to as the freezing robot problem \cite{Trautman2010}. Planning approaches considering reciprocal reactions are categorized as interaction-aware and perform prediction and planning together in one step. They promise to generate less conservative, more human-like, and progressive trajectories by influencing the other vehicles' behaviors to a certain extent with the knowledge that they will react to the ego vehicle. With this knowledge, interaction-aware approaches can also better avoid collisions, increasing the safety of the autonomous system.

Interaction-aware approaches have mainly been proposed for traffic scenarios like lane changes \cite{Schmidt2019}, ramp merges \cite{LeCleach2022}, or crosswalks \cite{FridovichKeil2020, Crosato2023}. They employ various methods, to name a few: multi-agent planning with a joint cost function, \aclp{POMDP}, reinforcement learning, and game-theoretical concepts. The latter provides concepts for non-cooperative behaviors and thus is especially fitting for autonomous racing with strategies like overtaking, blocking, and faking. Furthermore, game-theoretic concepts require assumptions about the cost functions that govern the players' decisions. In racing, the players share the same objective to finish ahead of the opponents, whereas traffic scenarios comprise a wide range of often unknown objectives.

With this contribution, we thoroughly analyze one of the game-theoretical approaches called \ac{iLQGame} for its suitability for trajectory planning in competitive racing scenarios. We focus on a straight race track section to systematically assess interaction awareness, competitiveness, and safety.

\subsection{Related Work}
\label{sec:related_work}
Most game-theoretic planning approaches are concerned with finding trajectories that fulfill the requirements of a Nash equilibrium. At a Nash equilibrium, no player is incentivized to alter its strategy unilaterally. Depending on the information structure of the formulated game, one obtains either the open-loop or the feedback solution. For an open-loop solution, each player must commit to a sequence of actions at the beginning of the game, directly defining the trajectories. For a feedback solution, the players commit to strategies that allow them to react to the current state in each stage of the game. A more detailed introduction to these two concepts will follow in Section \ref{sec:problem_formulation}. Both equilibrium concepts have been used in game-theoretic trajectory planning approaches, which we categorize into the following groups:
\subsubsection{Offline policy generation}
Methods of this category perform extensive offline computations to determine optimal policies. These policies can, e.g., be stored in lookup tables and applied efficiently online. Fisac et al. \cite{Fisac2019} discretize the combined state space of all players and determine the strategy for a feedback Stackelberg equilibrium via dynamic programming. This approach suffers from the curse of dimensionality, so only a few players and coarse discretizations are possible.
For racing, Bhargav et al. \cite{Bhargav2021} do not solve for equilibria but determine policies with a high probability of successful overtaking maneuvers for different race track positions. In extensive form, Zheng et al. \cite{Zheng2022} formulate racing as a two-player zero-sum game and determine the optimal strategy via \ac{CFR} minimization.

\subsubsection{Sampling-based}
Liniger and Lygeros \cite{Liniger2020} sample trajectory candidates for two players on a race track and formulate bi-matrix games. Nash and Stackelberg equilibria in the bi-matrices covering all possible trajectory combinations provide open-loop solutions. Like other methods that yield open-loop solutions, online re-planning with a moving horizon like in \ac{MPC} introduces feedback.

\subsubsection{\Ac{IBR}}
With \ac{IBR} approaches, the players optimize their trajectories alternately while keeping all other players' trajectories fixed. If this algorithm converges, no player is incentivized to alter its decision, making it a Nash equilibrium. Sensitivity-enhanced algorithms have been proposed in \cite{Wang2021, Wang.2019, Spica.2020} for drone and vehicle racing. Since the trajectories are optimized directly, the results are open-loop solutions.

\subsubsection{\Ac{DDP}}
\ac{DDP} \cite{MAYNE1966} is a trajectory optimization method that iteratively performs backward- and forward passes to refine the trajectory. During the backward pass, an incremental feedback law is generated based on second-order approximations of the cost and dynamics along a nominal trajectory. The forward pass updates the nominal trajectory based on the incremental feedback law. Using a first-order approximation of the dynamics results in \ac{iLQR} \cite{Li2004}.

Fridovich-Keil et al. \cite{FridovichKeil2020} transfer this iterative procedure to dynamic games. They approximate each player's cost function with a second-order tailor expansion and linearize the dynamics. The result is a linear quadratic game for which---like for time-discrete \acp{LQR}---analytic solutions exist \cite{Basar1999}. If this algorithm, called \ac{iLQGame}, converges, a Nash equilibrium to a local approximation of the game is found. A feature of \ac{iLQGame} is that it can provide both open-loop and feedback strategies for the players.
Similarly, Schwarting et al. \cite{Schwarting2021} solve a quadratic game in the backward pass to compute incremental feedback laws for the players. They plan in belief space, making it a multi-player variant of \ac{iLQG} control. 

Kavuncu et al. \cite{Kavuncu.2021} show that their used cost function constitutes a potential game, allowing reformulating the problem as a conventional \ac{OCP}. They use \ac{iLQR} to solve the \ac{OCP} and generate open-loop solutions.

\subsubsection{First-order optimality condition}
Le Cleac'h et al. \cite{LeCleach2022} solve a root-finding problem to fulfill the first-order optimality condition of a Nash equilibrium. The ALGames algorithm enforces constraints with an augmented Lagrangian method and yields open-loop solutions with reported superior computation times compared to \ac{iLQGame}.
Zhu and Borrelli \cite{Zhu.2023} develop a multi-player \ac{SQP} variant to find a Nash equilibrium as a solution to the \ac{KKT} conditions. As in \cite{LeCleach2022}, the algorithm finds open-loop solutions if it converges.

\subsection{Contributions and Outline}
In this paper, we apply the \ac{iLQGame} algorithm for planning in racing scenarios. The contributions are three-fold:
\begin{itemize}
	\item We analyze the applicability of \ac{iLQGame} in competitive racing scenarios. The evaluation is based on interaction awareness, competitiveness, and safety.
	\item We show that the cost parameterization adjusts the aggressiveness of the vehicles and can be used to distribute responsibilities among the players to avoid collisions.
	\item We compare the open-loop and the feedback solutions obtained by \ac{iLQGame} and show that they can result in fundamentally different behaviors.
\end{itemize}

Section~\ref{sec:problem_formulation} first introduces the game-theoretic preliminaries needed, followed by a detailed description of the \ac{iLQGame} algorithm in Section~\ref{sec:solving_game}. In Section~\ref{sec:racing_game}, we formulate the racing scenario as a dynamic game and provide details on the implementation. The results supporting the three contributions above are given in Section~\ref{sec:results}. Section~\ref{sec:dis_out} concludes whether \ac{iLQGame} is suitable for racing scenarios and provides an overview of future analyses.

\section{Game-Theoretic Preliminaries}
\label{sec:problem_formulation}
The discrete-time dynamics describing the propagation of the joint state $\bm{x}_k$ of a dynamic game with $N$ players is given by the time-variant non-linear function $\bm{f}_k$:
\begin{equation}
	\label{eq:game_nonlin_dynamics}
	\bm{x}_{k+1} = \bm{f}_k\left(\bm{x}_k, \bm{u}_k^1, \dots, \bm{u}_k^N \right)\text{.}
\end{equation}
We choose $\bm{x}_k \in \mathcal{X} = \mathbb{R}^n$ and $\bm{u}_k^i \in \mathcal{U}^i = \mathbb{R}^{m}$ for all players $i\in\mathcal{N}=\left\{1, 2, \dots, N\right\}$. In each of the $K$ stages player $i$ receives stage costs depending on the player's control inputs $\bm{u}_k^i \in \bm{u}^i=\left\{\bm{u}_0^i, \bm{u}_1^i, \dots, \bm{u}_{K-1}^i\right\}$ and the state $\bm{x}_k$. The sequence of states depends via \eqref{eq:game_nonlin_dynamics} on $\bm{u}^i$ and the control inputs of all other players, which is often expressed with the index $-i$. Given an initial state $\bm{x}_0$, the total cost of player $i$ can be written as:
\begin{align}
	\label{eq:game_nonlin_cost_input}
	J^i\left(\bm{x}_0, \bm{u}^i, \bm{u}^{-i}\right) = \sum_{k=0}^{K-1} g_k^i\left(\bm{x}_{k}, \bm{u}_k^i\right) +g_K^i\left(\bm{x}_{K}\right)
\end{align}
with the terminal cost $g_K^i\left(\bm{x}_{K}\right)$.

A strategy $\gamma^i=\left\{\bm{\gamma}_0^i(\cdot), \bm{\gamma}_1^i(\cdot), \dots, \bm{\gamma}_{K-1}^i(\cdot)\right\}$ of the strategy space $\Gamma^i=\left\{\Gamma_0^i, \Gamma_1^i, \dots, \Gamma_{K-1}^i\right\}$ determines the control inputs at each stage $k$, depending on the available information to player $i$. The cost functional $\eqref{eq:game_nonlin_cost_input}$ expressed with strategies is:
\begin{equation}
	\label{eq:game_nonlin_cost_strategy}
	J^i\left(\bm{x}_0, \gamma^i,\gamma^{-i}\right) = \sum_{k=0}^{K-1} g_k^i\left(\bm{x}_{k}, \bm{\gamma}_k^i(\cdot)\right)+g_K^i\left(\bm{x}_{K}\right)\text{.}
\end{equation}
Omitting the dependency on $\bm{x}_0$ for brevity in the following, an $N$-tuple of strategies $\left\{\gamma^{i*}\in\Gamma^i; i \in \mathcal{N}\right\}$ constitutes a Nash equilibrium if:
\begin{equation}
	\label{eq:nash_equilibrium}
	\forall i \in \mathcal{N}: J^i\left(\gamma^{i*}, \gamma^{-i*}\right) \le J^i\left(\gamma^{i}, \gamma^{-i*}\right)\text{.}
\end{equation}
In other words, no player can improve its outcome by unilaterally altering its strategy.

\subsection{Solution Concepts}
The domain and codomain of the functions in the strategy space depend on the information structure of the game \cite{Basar1999}. The two information structures we consider lead to the following two types of solution concepts:
\subsubsection{Open-loop solution} 
In the open-loop case, all players observe the initial state $\bm{x}_0$ and generate a sequence of control inputs in a single act. The strategy at stage $k$ is a constant function with $\bm{\gamma}_k^i(\cdot) \in \Gamma_k^i=\mathcal{U}^i$. A Nash equilibrium $\left\{\gamma^{i*}\in\Gamma^i; i \in \mathcal{N}\right\}$ therefore directly translates to the players' input sequences $\left\{\bm{u}^{i*} = \gamma^{i*}; i \in \mathcal{N}\right\}$. Beginning at $\bm{x}_0^*=\bm{x}_0$, the discrete dynamics \eqref{eq:game_nonlin_dynamics} provide the corresponding open-loop trajectory $\left\{\bm{x}_{k+1}^*; k\in\{0, 1, \dots, K-1\}\right\}$.

Another viewpoint considers the open-loop problem in discrete time as a static infinite game. There are infinite possible control input sequences of which each player must choose one at the first and only stage $k=0$ \cite{Basar1999}.

\subsubsection{Feedback solution}
In the feedback case, the players observe the current state $\bm{x}_k$ and are not bound to an initially announced sequence of control inputs. A feedback strategy $\gamma_k^i: \mathcal{X}\to\mathcal{U}^i$ maps the current state to a control input $\bm{u}_k^i$ so that the control inputs at a stage $k$ for a Nash equilibrium are: $\left\{\bm{u}_k^{i*} = \gamma_k^{i*}(\bm{x}_k); i \in \mathcal{N}\right\}$. For discretized state spaces, such strategies can be obtained via dynamic programming, also known as backward recursion in game theory. This involves proceeding backward from $k=K$ to $k=0$ and determining a Nash equilibrium for each static sub-game from stage $k$ to $k+1$.

\subsubsection*{ Connection to \acp{OCP}}
A game with $N=1$ simplifies to an \ac{OCP} as it involves only a single cost-functional $J^1$ for minimization. The trajectory for an initial state $\bm{x}_0$ obtained by using the discrete dynamics \eqref{eq:game_nonlin_dynamics} and applying the feedback solution coincides with the trajectory of the open-loop solution. This does not apply to the Nash equilibrium of non-zero-sum games with $N>1$, where the two solutions generally differ, even without disturbances or other unforeseen inputs. Starr and Ho \cite{Starr1969b} provide an illustrative example of this phenomenon and further details.

\section{Solving Discrete-Time Dynamic Games}
\label{sec:solving_game}
The \ac{iLQGame} algorithm in \cite{FridovichKeil2020} iteratively generates time-variant linear feedback laws for the players based on \ac{LQ} approximations of the game. Upon convergence, the approach yields a feedback solution for the last \ac{LQ} approximation. In the supplementary material to \cite{FridovichKeil2020}, the authors also demonstrate the capability of the algorithm to find the open-loop solution\footnote{\href{https://github.com/HJReachability/ilqgames}{https://github.com/HJReachability/ilqgames}}. In the following, we explain the steps of \ac{iLQGame}, which is summarized in Algorithm \ref{alg:iLQG}.
\begin{algorithm}[t]
	\small
	\caption{\ac{iLQGame}}
	\label{alg:iLQG}
	\begin{algorithmic}[1]
		\STATE \textbf{Input}: $\bm{x}_0$, initial guesses for $\hat{\bm{u}}^{i}$ (and $\hat{\bm{x}}$ using \eqref{eq:game_nonlin_dynamics}) \label{alg:initial_guess}
		\STATE \textbf{Output}: Nash equilibrium trajectory $\bm{u}^{i*}$ and $\bm{x}^{*}$
		\WHILE{not converged}
		\FOR{$k\in\{0, 1, \dots, K\}$}
		\STATE $\bm{A}_k, \bm{B}_k^i \leftarrow$ \textsc{Linearize}$(\hat{\bm{x}}_k, \hat{\bm{u}}_k^1, \dots, \hat{\bm{u}}_k^N)$ \label{alg:linearize}
		\STATE $\bm{Q}_k^i, \bm{q}_k^i, \bm{R}_k^{ii}, \bm{r}_k^{ii} \leftarrow$ \textsc{Quadratize}$(\hat{\bm{x}}_k, \hat{\bm{u}}_k^1, \dots, \hat{\bm{u}}_k^N)$\label{alg:quadratize}	
		\ENDFOR
		\STATE $\bm{K}_k^i, \bm{k}_k^i \leftarrow$ \textsc{SolveLQGame}$(\bm{A}_k, \bm{B}_k^i, \bm{Q}_k^i, \bm{q}_k^i, \bm{R}_k^{ii}, \bm{r}_k^{ii})$\hspace{-1cm}\label{alg:solve_lq_game}
		\FOR{$k\in\{0, 1, \dots, K\}$}
		\STATE $\hat{\bm{u}}_k^{i,\mathrm{new}} \leftarrow$ \textsc{UpdateInput}$(\hat{\bm{x}}_k^\mathrm{new}, \bm{K}_k^i, \bm{k}_k^i)$ \label{alg:update_input}
		\STATE $\hat{\bm{x}}_{k+1}^{\mathrm{new}} = \bm{f}_k\left(\hat{\bm{x}}_k^{\mathrm{new}}, \hat{\bm{u}}_k^{1,\mathrm{new}}, \dots, \hat{\bm{u}}_k^{N,\mathrm{new}} \right)$ \label{alg:forward_pass}
		\ENDFOR
		\STATE $\hat{\bm{u}}^i \leftarrow \hat{\bm{u}}^{i, \mathrm{new}}$, $\hat{\bm{x}} \leftarrow \hat{\bm{x}}^{\mathrm{new}}$
		\ENDWHILE
		\STATE$\bm{u}^{i*} \leftarrow \hat{\bm{u}}^i$, $\bm{x}^{*}\leftarrow \hat{\bm{x}}$
	\end{algorithmic}
\end{algorithm}
With an initial state $\bm{x}_0$ and an initial guess for each player's control input sequence $\hat{\bm{u}}^i$, the game dynamics \eqref{eq:game_nonlin_dynamics} yield the initial nominal trajectory $\hat{\bm{x}}$.

\subsubsection{Linearization of the dynamics}\label{sec:linearization}
A linearization along the nominal trajectory yields a linear time-discrete and time-variant state space model:
\begin{equation}
	\label{eq:game_linearized_dynamics}
	\begin{aligned}
		\Delta\bm{x}_{k+1} &= \bm{A}_k\Delta\bm{x}_{k} + \sum_{i=1}^{N}\bm{B}_k^i\Delta\bm{u}_{k}^i \text{ with }\\ \Delta\bm{x}_{k} &= \bm{x}_{k} - \hat{\bm{x}}_k \text{ and } \Delta\bm{u}_{k}^i = \bm{u}_{k} - \hat{\bm{u}}_k^i\text{.}
	\end{aligned}
\end{equation}
\subsubsection{Quadratic approximation of the cost function}\label{sec:quadratization}
As in \ac{iLQR}, the players' cost functions are approximated quadratically:
\begin{equation}
	\label{eq:game_quadratized_cost}
	\begin{split}
		g_k^i&\left(\Delta\bm{x}_k, \Delta\bm{u}_k^1, \dots, \Delta\bm{u}_k^N\right)\thickapprox g_k^i\left(\hat{\bm{x}}_k, \hat{\bm{u}}_k^1, \dots, \hat{\bm{u}}_k^N\right) + \\&\qquad\underbrace{\left(\nabla_\mathbf{x}g_k^i|_{\hat{\bm{x}}_k,\hat{\bm{u}}_k^1,\dots,\hat{\bm{u}}_k^N}\right)^\top}_{\bm{q}_k^i} \Delta\bm{x}_{k} + 
		\\&\qquad\frac{1}{2}\Delta\bm{x}_{k}^\top\underbrace{\left(\nabla_\mathbf{x}^2 g_k^i|_{\hat{\bm{x}}_k,\hat{\bm{u}}_k^1,\dots,\hat{\bm{u}}_k^N}\right)}_{\bm{Q}_k^i}\Delta\bm{x}_{k} + 
		\\&\qquad\underbrace{\left(\nabla_{\mathbf{u}^i} g_k^i|_{\hat{\bm{x}}_k,\hat{\bm{u}}_k^1,\dots,\hat{\bm{u}}_k^N}\right)^\top}_{\bm{r}_k^{ii}} \Delta\bm{u}_{k}^i +
		\\&\qquad\frac{1}{2}\Delta\bm{u}_{k}^{i\top}\underbrace{\left(\nabla_{\mathbf{u}^i}^2 g_k^i|_{\hat{\bm{x}}_k,\hat{\bm{u}}_k^1,\dots,\hat{\bm{u}}_k^N}\right)}_{\bm{R}_k^{ii}}\Delta\bm{u}_{k}^i\text{.}
	\end{split}
\end{equation}
The approximated cost functions can additionally have mixed second-order terms, which we omit here for brevity since they do not appear in our cost functions. Using $\bm{q}_k^i$, $\bm{Q}_k^i$, $\bm{r}_k^i$, and $\bm{R}_k^i$ and omitting the constant term, the total cost for player $i$ can be rewritten as:
\begin{equation}
	\label{eq:total_quad_cost}
	\begin{split}
		J^i \propto &\frac{1}{2}\sum_{k=0}^{K-1}\Biggl[\left(\Delta\bm{x}_k^\top \bm{Q}_k^i + 2\bm{q}_k^{i\top}\right)\Delta\bm{x}_k \Biggr.+\\ 
		&\qquad\quad\Biggl.\left(\Delta\bm{u}_k^{i\top} \bm{R}_k^{ii} + 2\bm{r}^{ii\top}\right)\Delta\bm{u}_k^i\Biggr] + \\
		& \frac{1}{2}\Delta\bm{x}_K^\top\bm{Q}_K^i\Delta\bm{x}_K + \bm{q}_K^{i\top}\Delta\bm{x}_K\text{.}
	\end{split}
\end{equation}
The player costs \eqref{eq:total_quad_cost} and the linear dynamics \eqref{eq:game_linearized_dynamics} constitute a \ac{LQ} game that approximates the game locally around the current nominal trajectory $\hat{\bm{x}}$.

\subsubsection{Solving the \ac{LQ} game}
For a \ac{LQ} game, a unique analytical solution for a Nash equilibrium exists. Like for \acp{LQR}, a feedback strategy has the linear affine form $\gamma_k^{i*}(\Delta\bm{x}_k) = -\bm{K}_k^i\Delta\bm{x}_k - \bm{k}_k^i$ \cite{Basar1999}. The elements in the matrices $\bm{K}_k^i$ and vectors $\bm{k}_k^i$ are obtained by solving the following systems of linear equations \cite{Basar1999, FridovichKeil2020}:
\begin{subequations}
	\label{eq:feedback_sol}
	\begin{align}
		&\begin{aligned}
			\left(\bm{R}_k^{ii} + \bm{B}_k^{i\top}\right. &\left.\bm{P}_{k+1}^i\bm{B}_k^i\right)\bm{K}_k^i + \bm{B}_k^{i\top}\bm{P}_{k+1}^i\cdot \\
			&\sum_{j=1, j\neq i}^{N}\bm{B}_k^j\bm{K}_k^j =\bm{B}_k^{i\top}\bm{P}_{k+1}^i\bm{A}_k
		\end{aligned}\label{eq:lin_sys_K}\\
		&\begin{aligned}\left(\bm{R}_k^{ii} + \bm{B}_k^{i\top}\right. &\left. \bm{P}_{k+1}^i\bm{B}_k^i\right)\bm{k}_k^i + \bm{B}_k^{i\top}\bm{P}_{k+1}^i\cdot \\
			&\sum_{j=1, j\neq i}^{N}\bm{B}_k^j\bm{k}_k^j =\bm{B}_k^{i\top}\bm{p}_{k+1}^i + \bm{r}_k^{ii}\text{.}
		\end{aligned}
	\end{align}
\end{subequations}
The recursion given in Appendix \ref{sec:recursion_feedback} provides the matrices $\bm{P}_k^i$ and vectors $\bm{p}_k^i$. The connection to \acp{LQR} becomes evident for $N=1$ and $\bm{q}_k = \bm{r}_k = 0$. In this case \eqref{eq:coupled_riccati} simplifies to the well known \emph{difference Riccati equation}, when substituting in \eqref{eq:lin_sys_K} and dropping index $i$:
\begin{equation}
	\begin{aligned}
		\bm{P}_k = \bm{Q}_k + &\bm{A}_k^\top\bm{P}_{k+1}\bm{A}_k - \left(\bm{A}_k^\top\bm{P}_{k+1}\bm{B}_k\right)\cdot\\&\left(\bm{R}_k+\bm{B}_k^\top\bm{P}_{k+1}\bm{B}_k\right)^{-1}\left(\bm{B}_k^\top\bm{P}_{k+1}\bm{A}_k\right)\text{.}
	\end{aligned}
\end{equation}

For the open-loop Nash equilibrium, the strategy $\gamma_k^{i*}(\cdot) = - \bm{k}_k^i$ does not depend on the current state and can be obtained with \cite{Basar1999, FridovichKeil2020}:
\begin{subequations}
	\begin{alignat}{2}
		&\bm{k}_k^{i} = -\bm{R}_k^{ii^{-1}}\left[\bm{B}_k^{i\top}\left(\bm{M}_{k+1}^i\Delta\bm{x}_{k+1} + \bm{m}_{k+1}^i\right) + \bm{r}_k^{ii}\right]\\
		&\begin{aligned}
			\Delta\bm{x}_{k+1} = \bm{\Lambda}_k^{-1}&\Biggl[\bm{A}_k\Delta\bm{x}_k - \Biggr. \\ &\Biggl.\sum_{j=1}^{N}\bm{B}_k^j\bm{R}_k^{jj^{-1}}\left(\bm{B}_k^{j\top}\bm{m}_{k+1}^j+\bm{r}_k^{jj}\right)\Biggr]\text{.}
		\end{aligned}
	\end{alignat}
\end{subequations}
The recursions to obtain $\bm{M}_{k+1}^i$ and $\bm{m}_{k+1}^i$ are given in Appendix \ref{sec:recursion_open_loop}. Both recursions for the feedback and the open-loop solution proceed backward from $K$ to $0$, so this step is called the backward pass.

\subsubsection{Update trajectory}
The forward pass updates the control inputs and the nominal trajectory according to the strategies generated in the backward pass. With the linearization of the dynamics and quadratic approximation of the cost function, the calculated control inputs apply additively to the nominal control inputs $\hat{\bm{u}}^i$ of the previous iteration. In the feedback case, $\bm{K}_k^i$ is multiplied by the deviation of the state from the previous iteration, and in the open-loop case, $\bm{K}_k^i$ is set to $\bm{0}$. Beginning with $\hat{\bm{x}}_0^{\mathrm{new}} = \hat{\bm{x}}_0 = \bm{x}_0$ the forward pass proceeds using \eqref{eq:game_nonlin_dynamics}:\\
For $k$ from $1$ to $K$:
\begin{subequations}
	\begin{alignat}{2}
		\hat{\bm{u}}_k^{i,\mathrm{new}} &= \hat{\bm{u}}_k^i - \bm{K}_k^i \left(\hat{\bm{x}}_k^{\mathrm{new}} - \hat{\bm{x}}_k\right) - \eta\bm{k}_k^i \\ \hat{\bm{x}}_{k+1}^{\mathrm{new}}&=\bm{f}_k\left(\hat{\bm{x}}_k^{\mathrm{new}}, \hat{\bm{u}}_k^{1, \mathrm{new}}, \dots, \hat{\bm{u}}_k^{N, \mathrm{new}}\right)\text{.}
	\end{alignat}
\end{subequations}
The scalar parameter $0<\eta\le 1$ is the step size and is usually chosen much smaller than $1$. It accounts for large deviations from the nominal trajectory where the \ac{LQ} approximation does not hold. With the new trajectory as the nominal trajectory, linearization, quadratic approximation, backward pass, and forward pass repeat until the algorithm converges.

\section{Racing Game}
\label{sec:racing_game}
\subsection{Vehicle Model and Game Dynamics}
As in \cite{Rowold.2023}, we model each player using a point mass with the state including the progress $s$, velocity $V$, lateral displacement $n$, relative orientation $\chi$ towards the track's center line with curvature $\kappa(s)$, and the longitudinal and lateral accelerations $a_\mathrm{x}$ and $a_\mathrm{y}$. The control input vector includes the jerks in longitudinal and lateral directions: $\bm{u}^\top = \begin{bmatrix}
	j_\mathrm{x} & j_\mathrm{y}
\end{bmatrix}$. The time-continuous nonlinear dynamics of player $i$ are given by:
\begin{equation}
	\label{eq:dynamics}
	\dot{\mathbf{x}}^i = \begin{bmatrix}
		\dot{s}^i\\
		\dot{V}^i\\
		\dot{n}^i\\
		\dot{\chi}^i\\
		\dot{a}_\mathrm{x}^i\\
		\dot{a}_\mathrm{y}^i
	\end{bmatrix}=\tilde{\mathbf{f}}^i(\mathbf{x}^i, \mathbf{u}^i) = \begin{bmatrix}
		\frac{V^i\cos(\chi^i)}{1-n^i\kappa(s^i)}\\
		a_\mathrm{x}^i\\
		V^i\sin(\chi^i) \\
		\frac{a_\mathrm{y}^i}{V^i}-\kappa(s^i)\frac{V^i\cos(\chi^i)}{1-n^i\kappa(s^i)}\\
		j_\mathrm{x}^i\\
		j_\mathrm{y}^i
	\end{bmatrix}\text{.}
\end{equation}
The joint state vector of the game is a concatenation of $N$ player state vectors:
\begin{equation}
	\dot{\bm{x}} = \begin{bmatrix}
		\dot{\bm{x}}^1\\
		\vdots\\
		\dot{\bm{x}}^N
	\end{bmatrix} = \begin{bmatrix}
		\tilde{\mathbf{f}}^1(\mathbf{x}, \mathbf{u}^i)\\
		\vdots\\
		\tilde{\mathbf{f}}^N(\mathbf{x}, \mathbf{u}^N)
	\end{bmatrix} = \tilde{\bm{f}}(\bm{x}, \bm{u}^1, \dots, \bm{u}^N)\text{.}
\end{equation}
The discretization and linearization in step~\ref{sec:linearization} are performed simultaneously with a forward Euler method:
\begin{subequations}
	\begin{align}
		\bm{A}_k &= \mathbb{I} + \Delta t\left.\frac{\partial\tilde{\bm{f}}}{\partial\bm{x}}\right\rvert_{\hat{\bm{x}}_k,\hat{\bm{u}}_k^1, \dots, \hat{\bm{u}}_k^N} \text{\quad and}\\ 
		\bm{B}_k^i &= \Delta t\left.\frac{\partial\tilde{\bm{f}}}{\partial\bm{u}^i}\right\rvert_{\hat{\bm{x}}_k,\hat{\bm{u}}_k^1, \dots, \hat{\bm{u}}_k^N}
	\end{align}
\end{subequations}
with the time discretization step size $\Delta t$ and the identity matrix $\mathbb{I}$.

Racing cars often operate at the handling limits, so it is essential to constrain the accelerations according to vehicle specifications. Similar to \cite{Rowold.2023}, we approximate the velocity-dependent gg-diagrams by diamonds with a maximum positive acceleration $a_\mathrm{x}\le a_\mathrm{x, max}(V)$ and a maximum combined acceleration specified by $a_\mathrm{x, min}(V)$ and $a_\mathrm{y, max}(V)$:
\begin{equation}
	\left(\frac{a_\mathrm{x}}{a_\mathrm{x, min}(V)}\right)^2 + \left(\frac{a_\mathrm{y}}{a_\mathrm{y, max}(V)}\right)^2 \le 1\text{.}
\end{equation}
The following cost function realizes the acceleration and other constraints.

\subsection{Cost Function}
\label{sec:cost_function}
\ac{LQR} related approaches naturally do not consider state and input constraints. Chen et al. \cite{Chen2017} realize constraints in \ac{iLQR} through the cost function and introduce barrier functions. For \ac{iLQGame}, quadratic cost terms for constraint violations as in \cite{FridovichKeil2020} achieve good results regarding convergence and robustness. The stage costs of player $i$ in our racing game, including the constraints, are:
\begin{subequations}
	\label{eq:stage_cost}
	\begin{alignat}{2}
		g&_k^i =\bm{u}_k^{i\top}\bm{R}^i\bm{u}_k^i+\label{eq:regularization}\\
		&\sum_{j=1, j\neq i}^{N}c_\mathrm{c}^ie^{1-\left(\frac{s_k^i -s_k^j}{l_\mathrm{veh}}\right)^2 - \left(\frac{n_k^i -n_k^j}{w_\mathrm{veh}}\right)^2}\label{eq:collision_cost} +\\
		&\bm{1}\left\{n_k^i\ge w_\mathrm{tr, l/r}(s_k^i) \right\}c_\mathrm{w}^i\left(n_k^i-w_\mathrm{tr, l/r}(s_k^i)\right)^2 + \label{eq:track_cost}\\
		&\bm{1}\left\{a_{\mathrm{x},k}^i\ge a_\mathrm{x, max}(V_k^i)\right\}c_{a_\mathrm{x}}^i\left(a_{\mathrm{x}, k}^i- a_\mathrm{x, max}(V_k^i)\right)^2 + \label{eq:ax_max_constr}\\
		&\begin{aligned}
			\bm{1}&\left\{\left(\frac{a_\mathrm{x}}{a_\mathrm{x, min}(V)}\right)^2 + \left(\frac{a_\mathrm{y}}{a_\mathrm{y, max}(V)}\right)^2 \ge 1\right\}c_a^i\cdot\\ &\left(\left(\frac{a_\mathrm{x}}{a_\mathrm{x, min}(V)}\right)^2 + \left(\frac{a_\mathrm{y}}{a_\mathrm{y, max}(V)}\right)^2 - 1\right)^2\text{.}
		\end{aligned}\label{eq:acc_constr}
	\end{alignat}
\end{subequations}
As in \cite{FridovichKeil2020}, the operator $\bm{1}\{\cdot\}$ becomes $1$ if the condition holds, and $0$ otherwise. The term \eqref{eq:regularization} regularizes the jerk as in \cite{Rowold.2023}. \eqref{eq:collision_cost} introduces a coupling between players by penalizing collisions. As player $i$ and $j$ come closer, the term increases with longitudinal and lateral distances weighted differently through the vehicle length $l_\mathrm{veh}$ and width $w_\mathrm{veh}$. The factor $c_\mathrm{c}^i$ weights the collision cost of player $i$. The other stage cost terms implement constraints with the weights $c_\mathrm{w}^i$, $c_{a_\mathrm{x}}^i$, $c_a^i$. \eqref{eq:track_cost} enforces the track boundaries with the track widths to the left and right $w_\mathrm{tr, l/r}$. \eqref{eq:ax_max_constr} and \eqref{eq:acc_constr} penalize violations of the previously mentioned acceleration constraints. 

The only coupling term \eqref{eq:collision_cost} penalizes collisions that can be avoided cooperatively so that the stage cost \eqref{eq:stage_cost} alone does not provide an incentive to race competitively. The following terminal cost term includes the progress of player $i$ and of all other players as in \cite{Wang2021} and introduces a competitive coupling between the players:
\begin{equation}
	\label{eq:end_cost}
	g_K^i = -s_K^i + c_\mathrm{g}^i \sum_{j=1, j\neq i}^{N}s_K^j\text{.}
\end{equation}
The first term with a negative sign penalizes little progress of player $i$. The second term with a positive sign and weight $c_\mathrm{g}$ penalizes the progress of other players, which provides an incentive to overtake in a trailing position and to defend a leading position, e.g., with a blocking maneuver. It should be noted that the gradients and Hessians of both the stage and terminal cost functions do not exhibit mixed second-order terms, so the form in \eqref{eq:game_quadratized_cost} can be used.

\subsection{Moving Horizon Implementation}
\label{sec:moving_horizon}
\ac{iLQGame} iteratively creates \ac{LQ} approximations of the game that hold only for small deviations from the nominal trajectory. With this local approximation, the solution to which \ac{iLQGame} converges depends on the initial guess of the players' input sequences in Line~\ref{alg:initial_guess} of Algorithm~\ref{alg:iLQG}. This initial guess can be decisive regarding behavioral decisions, e.g., overtaking on the left or right side and blocking or making way. 
Planning approaches usually operate with a moving horizon. In this way, they consider the receding initial state and the constantly changing environment with agents that do not behave as assumed, which occurs with both sequential and interaction-aware approaches. In a moving horizon implementation, the players' control inputs can be initialized with the solution of the previous planning step. Such an initialization reduces the number of required iterations and prevents the solution from alternating between different behavior classes. The vehicle commits to a particular behavior over consecutive planning steps.

While approximations with \ac{LQ} games exhibit unique Nash equilibria, the original game may not exhibit any Nash equilibrium or only ones unreachable for \ac{iLQGame} beginning with the initial guess. Such a situation can lead to oscillations over the iterations of \ac{iLQGame}, preventing the algorithm from converging. The decision-making process of a human racing driver attempting to pass another competitive vehicle can illustrate these oscillations: If the leading vehicle drives on the left, the driver's instinct is to overtake on the right. However, the driver anticipates a blocking maneuver and decides to stay left. The driver assumes that the leading vehicle knows about this reasoning and thus anticipates the leading vehicle to stay left once again. This results in a cyclic reasoning.
Considering the oscillations without convergence, limiting the number of iterations, and terminating the algorithm before convergence are crucial for an online application to meet computation time requirements. Feasibility of intermediate iterations is still ensured, as the forward pass uses the game dynamics \eqref{eq:dynamics}. The application of the algorithm with the results in the following Section~\ref{sec:results} has shown that non-converging planning steps only occur sporadically and seem insignificant with a moving horizon implementation. Regular operation is restored within a few planning steps.

\section{Results and Discussion}
\label{sec:results}
The following simulation results with a moving horizon implementation provide a foundation for evaluating the applicability of \ac{iLQGame} in competitive racing. We consider a head-to-head scenario similar to the motivating example in Fig.~\ref{fig:yielding_behavior} and call the leading vehicle the ego vehicle (player $1$) and the trailing vehicle the opponent (player $2$). We mainly analyze the defending behavior, i.e., the ego vehicle's behavior, but Section~\ref{sec:open_loop_vs_feedback} also allows conclusions about scenarios in which the ego vehicle is trailing with the game theoretical planning approach. In the following scenarios, the ego vehicle starts with its maximum speed of \SI{30}{\meter\per\second} \SI{50}{\meter} in front of the opponent. The opponent's maximum and initial speed is \SI{40}{\meter\per\second}.

A sequential planning approach is used as a reference. It is based on \ac{iLQGame} algorithm for one player, which is \ac{iLQR} and uses the same cost function with \eqref{eq:stage_cost} and \eqref{eq:end_cost} for maximized comparability. With $N=1$, the other players' positions in \eqref{eq:collision_cost} and \eqref{eq:end_cost} are not part of the state space but instead predicted using a constant velocity assumption on a straight path. Unless specified otherwise, all players share the same cost parameterization.

Section~\ref{sec:interaction_awareness_parameter} demonstrates the interaction awareness of the ego vehicle when using \ac{iLQGame} and illustrates the importance and meaning of the players' cost parameterization, exemplified with the collision cost term. The simulations in Section~\ref{sec:open_loop_vs_feedback} compare the behaviors for all combinations of planning approaches in the head-to-head scenario and highlight the difference between open-loop and feedback solutions.

\subsection{Interaction Awareness and Cost Parameterization}
\label{sec:interaction_awareness_parameter}
\begin{figure}[t]
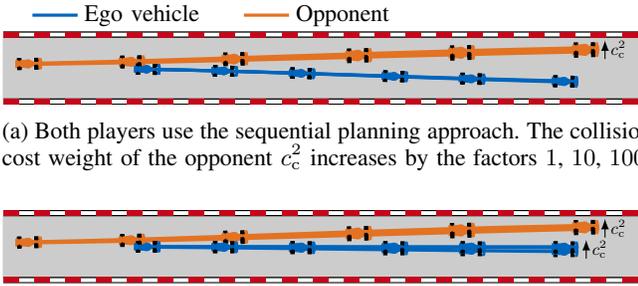

	\begin{subfigure}[t]{\linewidth}
		\small
		\centering
		\def\axiswidth{8.5cm}
		\input{figures/cc2_100.00_n1_0.00_n2_0.50_p1_iLQR_open_p2_iLQR_open_eta_0.1}
		\subcaption{Both players use the sequential planning approach. The collision cost weight of the opponent $c_\mathrm{c}^2$ increases by the factors $1$, $10$, $100$.}
		\label{fig:both_ilqr}
	\end{subfigure}\vspace{0.5cm}
	\begin{subfigure}[t]{\linewidth}
		\small
		\centering
		\def\axiswidth{8.5cm}
		\input{figures/cc2_100.00_n1_0.00_n2_0.50_p1_iQG_open_p2_iLQR_open_eta_0.1}
		\subcaption{The ego vehicle uses \ac{iLQGame} with an open-loop strategy, and the opponent uses the sequential planning approach. The collision cost weight of the opponent $c_\mathrm{c}^2$ increases by the factors $1$, $10$, $100$.}
		\label{fig:collision_cost}
	\end{subfigure}
	\caption{Interaction awareness and dependency on the cost parameterization.}
	\label{fig:single_steps}
\end{figure}
Fig.~\ref{fig:single_steps} shows the paths for two combinations of planning approaches and various collision cost weights in the opponent's cost function. In Fig.~\ref{fig:both_ilqr}, both players use the sequential planning approach and swerve outside due to the straight predictions with constant velocities. As the opponent's collision cost weight increases, the left-swerving maneuver's amplitude increases. However, the ego vehicle is unaffected by the cost increase as $c_\mathrm{c}^2$ of the opponent does not enter the \ac{iLQR} algorithm. The ego vehicle in Fig.~\ref{fig:collision_cost} uses the interaction-aware planning approach with \ac{iLQGame} and an open-loop strategy. The ego vehicle still makes way but swerves right less compared to Fig.~\ref{fig:both_ilqr}. As the collision cost weight of the opponent increases, the evasive movement of the ego vehicle decreases, and it almost does not react for $c_\mathrm{c}^2=100c_\mathrm{c}^1$. This demonstrates the ego vehicle's knowledge about the opponent's collision avoidance, which results in a less conservative, interaction-aware behavior.

In addition to demonstrating interaction awareness, an unequal cost parameterization can serve a purpose in distributing responsibilities to avoid crashes and implementing race rules. In the given scenario, higher collision costs imply that the opponent is more responsible for avoiding collisions due to its trailing position with a better scene overview. Simultaneously, the ego vehicle's aggressiveness increases. A similar effect occurs as the weight $c_\mathrm{g}^1$ increases, penalizing the opponent's progress and providing an incentive to slow it down.

\subsection{Open-Loop vs. Feedback}
\label{sec:open_loop_vs_feedback}
In the following simulations, the ego vehicle's and opponent's planning algorithms vary. Both vehicles are randomly placed on different lateral positions in the upper track half with the only constraint that the opponent is closer to the wall than the ego vehicle. This allows the ego vehicle to slow down or even block the opponent. For each planning approach combination, 260 simulations are performed once with $\nicefrac{c_\mathrm{c}^2}{c_\mathrm{c}^1}=1$ and once with $\nicefrac{c_\mathrm{c}^2}{c_\mathrm{c}^1}=10$. The time it takes the opponent to overtake and reach a longitudinal gap of \SI{20}{\meter} serves as a performance measure, and the collision probability as a safety measure. The mean overtaking times and collision probabilities for $\nicefrac{c_\mathrm{c}^2}{c_\mathrm{c}^1} = 1$ and $\nicefrac{c_\mathrm{c}^2}{c_\mathrm{c}^1} = 10$ are shown in Tables~\ref{tab:table1} and \ref{tab:table10}, respectively.
\renewcommand{\arraystretch}{1.2}
\begin{table}[t]
	\centering
	\caption{Mean overtaking times and collision probabilities for $c_\mathrm{c}^2 = c_\mathrm{c}^1$.}
	\label{tab:table1}
		\begin{tabular}{|c"*{4}{>{\centering\arraybackslash}p{1.7cm}|}}
		\cline{1-4}
		\diagbox[]{Ego}{Opp.} & Sequential & \begin{tabular}{@{}c@{}}\ac{iLQGame} \\ (open-loop)\end{tabular} & \begin{tabular}{@{}c@{}}\ac{iLQGame} \\ (feedback)\end{tabular} \\\thickhline
		\multirow{2}{*}{Sequential} & \SI{8.20}{\second} & \SI{7.08}{\second} & \SI{6.88}{\second} \\
		 & \SI{10.77}{\percent} & \SI{9.62}{\percent} & \SI{12.74}{\percent} \\
		\cline{1-4}
		\multirow{2}{*}{\begin{tabular}{@{}c@{}}\ac{iLQGame} \\ (open-loop)\end{tabular}} & \SI{8.60}{\second} & \SI{7.86}{\second} & \SI{7.20}{\second} \\
		 & \SI{0.38}{\percent} & \SI{0.00}{\percent} & \SI{10.55}{\percent} \\
		\cline{1-4}
		\multirow{2}{*}{\begin{tabular}{@{}c@{}}\ac{iLQGame} \\ (feedback)\end{tabular}} & \SI{8.81}{\second} & \SI{7.62}{\second} & \SI{7.14}{\second} \\
		 & \SI{0.39}{\percent} & \SI{1.94}{\percent} & \SI{0.00}{\percent} \\
		\cline{1-4}
	\end{tabular}

\end{table}
\begin{table}[t]
	\centering
	\caption{Mean overtaking times and collision probabilities for $c_\mathrm{c}^2 =10 c_\mathrm{c}^1$.}
	\label{tab:table10}
		\begin{tabular}{|c"*{4}{>{\centering\arraybackslash}p{1.7cm}|}}
		\cline{1-4}
		\diagbox[]{Ego}{Opp.} & Sequential & \begin{tabular}{@{}c@{}}\ac{iLQGame} \\ (open-loop)\end{tabular} & \begin{tabular}{@{}c@{}}\ac{iLQGame} \\ (feedback)\end{tabular} \\\thickhline
		\multirow{2}{*}{Sequential} & \SI{9.30}{\second} & \SI{7.32}{\second} & \SI{7.40}{\second} \\
		 & \SI{12.74}{\percent} & \SI{7.81}{\percent} & \SI{6.59}{\percent} \\
		\cline{1-4}
		\multirow{2}{*}{\begin{tabular}{@{}c@{}}\ac{iLQGame} \\ (open-loop)\end{tabular}} & \SI{10.90}{\second} & \SI{8.67}{\second} & \SI{9.70}{\second} \\
		 & \SI{0.39}{\percent} & \SI{0.00}{\percent} & \SI{0.39}{\percent} \\
		\cline{1-4}
		\multirow{2}{*}{\begin{tabular}{@{}c@{}}\ac{iLQGame} \\ (feedback)\end{tabular}} & \SI{14.42}{\second} & \SI{10.94}{\second} & \SI{11.28}{\second} \\
		 & \SI{6.72}{\percent} & \SI{0.00}{\percent} & \SI{0.00}{\percent} \\
		\cline{1-4}
	\end{tabular}

\end{table}

\begin{figure}[t]
	\begin{subfigure}[t]{\linewidth}
		\small
		\centering
		\def\axiswidth{7.0cm}
		\def\axisheight{3.0cm}
\begin{tikzpicture}

\definecolor{chocolate22711434}{RGB}{227,114,34}
\definecolor{darkcyan0101189}{RGB}{0,101,189}
\definecolor{darkgoldenrod1621730}{RGB}{162,173,0}
\definecolor{darkgray176}{RGB}{176,176,176}

\begin{axis}[
axis lines=left,
axis on top=true,
height=\axisheight,
legend cell align={left},
legend style={/tikz/every even column/.append style={column sep=0.5cm}},
legend style={fill opacity=0.0, draw opacity=1, text opacity=1, draw=none, fill=none},
scale only axis,
tick align=outside,
tick pos=left,
width=\axiswidth,
x grid style={darkgray176},
xlabel={Overtaking time in \si{\second}},
xmin=5.3, xmax=20.7,
xtick style={color=black},
y grid style={darkgray176},
ylabel={Frequency},
ymin=0, ymax=163.8,
ytick style={color=black}
]
\draw[draw=none,fill=darkcyan0101189,fill opacity=0.5] (axis cs:6,0) rectangle (axis cs:6.93333333333333,0);
\addlegendimage{ybar,ybar legend,draw=none,fill=darkcyan0101189,fill opacity=0.5}
\addlegendentry{Sequential}

\draw[draw=none,fill=darkcyan0101189,fill opacity=0.5] (axis cs:6.93333333333333,0) rectangle (axis cs:7.86666666666667,156);
\draw[draw=none,fill=darkcyan0101189,fill opacity=0.5] (axis cs:7.86666666666667,0) rectangle (axis cs:8.8,34);
\draw[draw=none,fill=darkcyan0101189,fill opacity=0.5] (axis cs:8.8,0) rectangle (axis cs:9.73333333333333,33);
\draw[draw=none,fill=darkcyan0101189,fill opacity=0.5] (axis cs:9.73333333333333,0) rectangle (axis cs:10.6666666666667,6);
\draw[draw=none,fill=darkcyan0101189,fill opacity=0.5] (axis cs:10.6666666666667,0) rectangle (axis cs:11.6,0);
\draw[draw=none,fill=darkcyan0101189,fill opacity=0.5] (axis cs:11.6,0) rectangle (axis cs:12.5333333333333,0);
\draw[draw=none,fill=darkcyan0101189,fill opacity=0.5] (axis cs:12.5333333333333,0) rectangle (axis cs:13.4666666666667,0);
\draw[draw=none,fill=darkcyan0101189,fill opacity=0.5] (axis cs:13.4666666666667,0) rectangle (axis cs:14.4,0);
\draw[draw=none,fill=darkcyan0101189,fill opacity=0.5] (axis cs:14.4,0) rectangle (axis cs:15.3333333333333,0);
\draw[draw=none,fill=darkcyan0101189,fill opacity=0.5] (axis cs:15.3333333333333,0) rectangle (axis cs:16.2666666666667,1);
\draw[draw=none,fill=darkcyan0101189,fill opacity=0.5] (axis cs:16.2666666666667,0) rectangle (axis cs:17.2,1);
\draw[draw=none,fill=darkcyan0101189,fill opacity=0.5] (axis cs:17.2,0) rectangle (axis cs:18.1333333333333,0);
\draw[draw=none,fill=darkcyan0101189,fill opacity=0.5] (axis cs:18.1333333333333,0) rectangle (axis cs:19.0666666666667,0);
\draw[draw=none,fill=darkcyan0101189,fill opacity=0.5] (axis cs:19.0666666666667,0) rectangle (axis cs:20,0);
\draw[draw=none,fill=darkgoldenrod1621730,fill opacity=0.5] (axis cs:6,0) rectangle (axis cs:6.93333333333333,0);
\addlegendimage{ybar,ybar legend,draw=none,fill=darkgoldenrod1621730,fill opacity=0.5}
\addlegendentry{\ac{iLQGame} (open-loop)}

\draw[draw=none,fill=darkgoldenrod1621730,fill opacity=0.5] (axis cs:6.93333333333333,0) rectangle (axis cs:7.86666666666667,134);
\draw[draw=none,fill=darkgoldenrod1621730,fill opacity=0.5] (axis cs:7.86666666666667,0) rectangle (axis cs:8.8,16);
\draw[draw=none,fill=darkgoldenrod1621730,fill opacity=0.5] (axis cs:8.8,0) rectangle (axis cs:9.73333333333333,17);
\draw[draw=none,fill=darkgoldenrod1621730,fill opacity=0.5] (axis cs:9.73333333333333,0) rectangle (axis cs:10.6666666666667,79);
\draw[draw=none,fill=darkgoldenrod1621730,fill opacity=0.5] (axis cs:10.6666666666667,0) rectangle (axis cs:11.6,10);
\draw[draw=none,fill=darkgoldenrod1621730,fill opacity=0.5] (axis cs:11.6,0) rectangle (axis cs:12.5333333333333,0);
\draw[draw=none,fill=darkgoldenrod1621730,fill opacity=0.5] (axis cs:12.5333333333333,0) rectangle (axis cs:13.4666666666667,0);
\draw[draw=none,fill=darkgoldenrod1621730,fill opacity=0.5] (axis cs:13.4666666666667,0) rectangle (axis cs:14.4,0);
\draw[draw=none,fill=darkgoldenrod1621730,fill opacity=0.5] (axis cs:14.4,0) rectangle (axis cs:15.3333333333333,1);
\draw[draw=none,fill=darkgoldenrod1621730,fill opacity=0.5] (axis cs:15.3333333333333,0) rectangle (axis cs:16.2666666666667,0);
\draw[draw=none,fill=darkgoldenrod1621730,fill opacity=0.5] (axis cs:16.2666666666667,0) rectangle (axis cs:17.2,0);
\draw[draw=none,fill=darkgoldenrod1621730,fill opacity=0.5] (axis cs:17.2,0) rectangle (axis cs:18.1333333333333,1);
\draw[draw=none,fill=darkgoldenrod1621730,fill opacity=0.5] (axis cs:18.1333333333333,0) rectangle (axis cs:19.0666666666667,0);
\draw[draw=none,fill=darkgoldenrod1621730,fill opacity=0.5] (axis cs:19.0666666666667,0) rectangle (axis cs:20,0);
\draw[draw=none,fill=chocolate22711434,fill opacity=0.5] (axis cs:6,0) rectangle (axis cs:6.93333333333333,0);
\addlegendimage{ybar,ybar legend,draw=none,fill=chocolate22711434,fill opacity=0.5}
\addlegendentry{\ac{iLQGame} (feedback)}

\draw[draw=none,fill=chocolate22711434,fill opacity=0.5] (axis cs:6.93333333333333,0) rectangle (axis cs:7.86666666666667,132);
\draw[draw=none,fill=chocolate22711434,fill opacity=0.5] (axis cs:7.86666666666667,0) rectangle (axis cs:8.8,12);
\draw[draw=none,fill=chocolate22711434,fill opacity=0.5] (axis cs:8.8,0) rectangle (axis cs:9.73333333333333,32);
\draw[draw=none,fill=chocolate22711434,fill opacity=0.5] (axis cs:9.73333333333333,0) rectangle (axis cs:10.6666666666667,45);
\draw[draw=none,fill=chocolate22711434,fill opacity=0.5] (axis cs:10.6666666666667,0) rectangle (axis cs:11.6,17);
\draw[draw=none,fill=chocolate22711434,fill opacity=0.5] (axis cs:11.6,0) rectangle (axis cs:12.5333333333333,4);
\draw[draw=none,fill=chocolate22711434,fill opacity=0.5] (axis cs:12.5333333333333,0) rectangle (axis cs:13.4666666666667,2);
\draw[draw=none,fill=chocolate22711434,fill opacity=0.5] (axis cs:13.4666666666667,0) rectangle (axis cs:14.4,1);
\draw[draw=none,fill=chocolate22711434,fill opacity=0.5] (axis cs:14.4,0) rectangle (axis cs:15.3333333333333,9);
\draw[draw=none,fill=chocolate22711434,fill opacity=0.5] (axis cs:15.3333333333333,0) rectangle (axis cs:16.2666666666667,2);
\draw[draw=none,fill=chocolate22711434,fill opacity=0.5] (axis cs:16.2666666666667,0) rectangle (axis cs:17.2,0);
\draw[draw=none,fill=chocolate22711434,fill opacity=0.5] (axis cs:17.2,0) rectangle (axis cs:18.1333333333333,0);
\draw[draw=none,fill=chocolate22711434,fill opacity=0.5] (axis cs:18.1333333333333,0) rectangle (axis cs:19.0666666666667,0);
\draw[draw=none,fill=chocolate22711434,fill opacity=0.5] (axis cs:19.0666666666667,0) rectangle (axis cs:20,0);
\end{axis}

\end{tikzpicture}
		\subcaption{Overtaking times for $\nicefrac{c_\mathrm{c}^2}{c_\mathrm{c}^1} = 1$.}
		\label{fig:hist1}
	\end{subfigure}\vspace{0.5cm}
	\begin{subfigure}[t]{\linewidth}
		\small
		\centering
		\def\axiswidth{7.0cm}
		\def\axisheight{3.0cm}
\begin{tikzpicture}

\definecolor{chocolate22711434}{RGB}{227,114,34}
\definecolor{darkcyan0101189}{RGB}{0,101,189}
\definecolor{darkgoldenrod1621730}{RGB}{162,173,0}
\definecolor{darkgray176}{RGB}{176,176,176}

\begin{axis}[
axis lines=left,
axis on top=true,
height=\axisheight,
legend cell align={left},
legend style={/tikz/every even column/.append style={column sep=0.5cm}},
legend style={fill opacity=0.0, draw opacity=1, text opacity=1, draw=none, fill=none},
scale only axis,
tick align=outside,
tick pos=left,
width=\axiswidth,
x grid style={darkgray176},
xlabel={Overtaking time in \si{\second}},
xmin=5.3, xmax=20.7,
xtick style={color=black},
y grid style={darkgray176},
ylabel={Frequency},
ymin=0, ymax=97.65,
ytick style={color=black}
]
\draw[draw=none,fill=darkcyan0101189,fill opacity=0.5] (axis cs:6,0) rectangle (axis cs:6.93333333333333,1);
\addlegendimage{ybar,ybar legend,draw=none,fill=darkcyan0101189,fill opacity=0.5}
\addlegendentry{Sequential}

\draw[draw=none,fill=darkcyan0101189,fill opacity=0.5] (axis cs:6.93333333333333,0) rectangle (axis cs:7.86666666666667,93);
\draw[draw=none,fill=darkcyan0101189,fill opacity=0.5] (axis cs:7.86666666666667,0) rectangle (axis cs:8.8,38);
\draw[draw=none,fill=darkcyan0101189,fill opacity=0.5] (axis cs:8.8,0) rectangle (axis cs:9.73333333333333,42);
\draw[draw=none,fill=darkcyan0101189,fill opacity=0.5] (axis cs:9.73333333333333,0) rectangle (axis cs:10.6666666666667,42);
\draw[draw=none,fill=darkcyan0101189,fill opacity=0.5] (axis cs:10.6666666666667,0) rectangle (axis cs:11.6,9);
\draw[draw=none,fill=darkcyan0101189,fill opacity=0.5] (axis cs:11.6,0) rectangle (axis cs:12.5333333333333,0);
\draw[draw=none,fill=darkcyan0101189,fill opacity=0.5] (axis cs:12.5333333333333,0) rectangle (axis cs:13.4666666666667,0);
\draw[draw=none,fill=darkcyan0101189,fill opacity=0.5] (axis cs:13.4666666666667,0) rectangle (axis cs:14.4,0);
\draw[draw=none,fill=darkcyan0101189,fill opacity=0.5] (axis cs:14.4,0) rectangle (axis cs:15.3333333333333,0);
\draw[draw=none,fill=darkcyan0101189,fill opacity=0.5] (axis cs:15.3333333333333,0) rectangle (axis cs:16.2666666666667,0);
\draw[draw=none,fill=darkcyan0101189,fill opacity=0.5] (axis cs:16.2666666666667,0) rectangle (axis cs:17.2,0);
\draw[draw=none,fill=darkcyan0101189,fill opacity=0.5] (axis cs:17.2,0) rectangle (axis cs:18.1333333333333,0);
\draw[draw=none,fill=darkcyan0101189,fill opacity=0.5] (axis cs:18.1333333333333,0) rectangle (axis cs:19.0666666666667,0);
\draw[draw=none,fill=darkcyan0101189,fill opacity=0.5] (axis cs:19.0666666666667,0) rectangle (axis cs:20,0);
\draw[draw=none,fill=darkgoldenrod1621730,fill opacity=0.5] (axis cs:6,0) rectangle (axis cs:6.93333333333333,0);
\addlegendimage{ybar,ybar legend,draw=none,fill=darkgoldenrod1621730,fill opacity=0.5}
\addlegendentry{\ac{iLQGame} (open-loop)}

\draw[draw=none,fill=darkgoldenrod1621730,fill opacity=0.5] (axis cs:6.93333333333333,0) rectangle (axis cs:7.86666666666667,57);
\draw[draw=none,fill=darkgoldenrod1621730,fill opacity=0.5] (axis cs:7.86666666666667,0) rectangle (axis cs:8.8,16);
\draw[draw=none,fill=darkgoldenrod1621730,fill opacity=0.5] (axis cs:8.8,0) rectangle (axis cs:9.73333333333333,6);
\draw[draw=none,fill=darkgoldenrod1621730,fill opacity=0.5] (axis cs:9.73333333333333,0) rectangle (axis cs:10.6666666666667,26);
\draw[draw=none,fill=darkgoldenrod1621730,fill opacity=0.5] (axis cs:10.6666666666667,0) rectangle (axis cs:11.6,47);
\draw[draw=none,fill=darkgoldenrod1621730,fill opacity=0.5] (axis cs:11.6,0) rectangle (axis cs:12.5333333333333,46);
\draw[draw=none,fill=darkgoldenrod1621730,fill opacity=0.5] (axis cs:12.5333333333333,0) rectangle (axis cs:13.4666666666667,16);
\draw[draw=none,fill=darkgoldenrod1621730,fill opacity=0.5] (axis cs:13.4666666666667,0) rectangle (axis cs:14.4,15);
\draw[draw=none,fill=darkgoldenrod1621730,fill opacity=0.5] (axis cs:14.4,0) rectangle (axis cs:15.3333333333333,16);
\draw[draw=none,fill=darkgoldenrod1621730,fill opacity=0.5] (axis cs:15.3333333333333,0) rectangle (axis cs:16.2666666666667,2);
\draw[draw=none,fill=darkgoldenrod1621730,fill opacity=0.5] (axis cs:16.2666666666667,0) rectangle (axis cs:17.2,1);
\draw[draw=none,fill=darkgoldenrod1621730,fill opacity=0.5] (axis cs:17.2,0) rectangle (axis cs:18.1333333333333,3);
\draw[draw=none,fill=darkgoldenrod1621730,fill opacity=0.5] (axis cs:18.1333333333333,0) rectangle (axis cs:19.0666666666667,1);
\draw[draw=none,fill=darkgoldenrod1621730,fill opacity=0.5] (axis cs:19.0666666666667,0) rectangle (axis cs:20,2);
\draw[draw=none,fill=chocolate22711434,fill opacity=0.5] (axis cs:6,0) rectangle (axis cs:6.93333333333333,0);
\addlegendimage{ybar,ybar legend,draw=none,fill=chocolate22711434,fill opacity=0.5}
\addlegendentry{\ac{iLQGame} (feedback)}

\draw[draw=none,fill=chocolate22711434,fill opacity=0.5] (axis cs:6.93333333333333,0) rectangle (axis cs:7.86666666666667,59);
\draw[draw=none,fill=chocolate22711434,fill opacity=0.5] (axis cs:7.86666666666667,0) rectangle (axis cs:8.8,7);
\draw[draw=none,fill=chocolate22711434,fill opacity=0.5] (axis cs:8.8,0) rectangle (axis cs:9.73333333333333,4);
\draw[draw=none,fill=chocolate22711434,fill opacity=0.5] (axis cs:9.73333333333333,0) rectangle (axis cs:10.6666666666667,6);
\draw[draw=none,fill=chocolate22711434,fill opacity=0.5] (axis cs:10.6666666666667,0) rectangle (axis cs:11.6,0);
\draw[draw=none,fill=chocolate22711434,fill opacity=0.5] (axis cs:11.6,0) rectangle (axis cs:12.5333333333333,0);
\draw[draw=none,fill=chocolate22711434,fill opacity=0.5] (axis cs:12.5333333333333,0) rectangle (axis cs:13.4666666666667,0);
\draw[draw=none,fill=chocolate22711434,fill opacity=0.5] (axis cs:13.4666666666667,0) rectangle (axis cs:14.4,0);
\draw[draw=none,fill=chocolate22711434,fill opacity=0.5] (axis cs:14.4,0) rectangle (axis cs:15.3333333333333,33);
\draw[draw=none,fill=chocolate22711434,fill opacity=0.5] (axis cs:15.3333333333333,0) rectangle (axis cs:16.2666666666667,18);
\draw[draw=none,fill=chocolate22711434,fill opacity=0.5] (axis cs:16.2666666666667,0) rectangle (axis cs:17.2,8);
\draw[draw=none,fill=chocolate22711434,fill opacity=0.5] (axis cs:17.2,0) rectangle (axis cs:18.1333333333333,5);
\draw[draw=none,fill=chocolate22711434,fill opacity=0.5] (axis cs:18.1333333333333,0) rectangle (axis cs:19.0666666666667,8);
\draw[draw=none,fill=chocolate22711434,fill opacity=0.5] (axis cs:19.0666666666667,0) rectangle (axis cs:20,10);
\end{axis}

\end{tikzpicture}
		\subcaption{Overtaking times for $\nicefrac{c_\mathrm{c}^2}{c_\mathrm{c}^1} = 10$.}
		\label{fig:hist2}
	\end{subfigure}\vspace{0.5cm}
	\caption{Frequencies of overtaking times when the opponent uses a sequential planning approach for different ego vehicle planning approaches.}
	\label{fig:hist}
\end{figure}
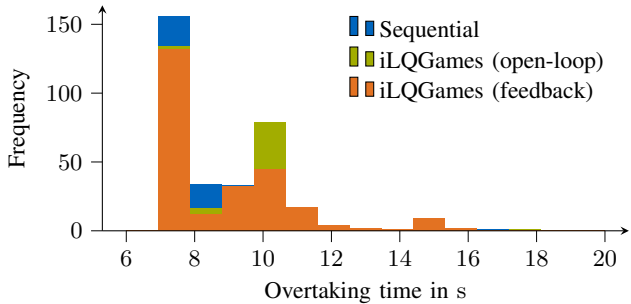
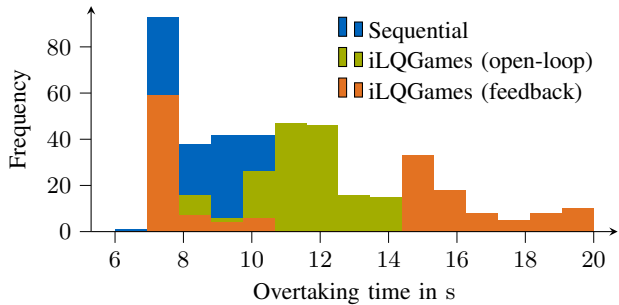

\begin{figure}[t]
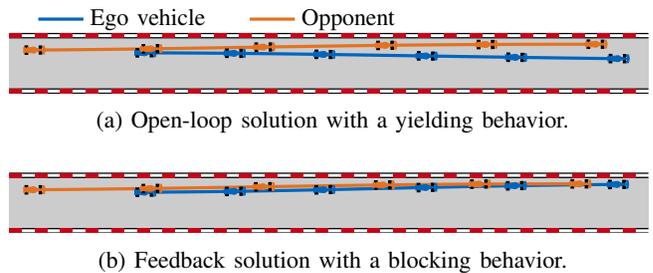

	\begin{subfigure}[t]{\linewidth}
		\small
		\centering
		\def\axiswidth{8.5cm}
		\input{figures/cc2_10.00_n1_2.50_n2_3.00_p1_iQG_open_p2_iLQR_open_eta_0.1}
		\subcaption{Open-loop solution with a yielding behavior.}
		\label{fig:iQG_open_iLQR_open}
	\end{subfigure}\vspace{0.5cm}
	\begin{subfigure}[t]{\linewidth}
		\small
		\centering
		\def\axiswidth{8.5cm}
		\input{figures/cc2_10.00_n1_2.50_n2_3.00_p1_iQG_closed_p2_iLQR_open_eta_0.1}
		\subcaption{Feedback solution with a blocking behavior.}
		\label{fig:iQG_closed_iLQR_open}
	\end{subfigure}
	\caption{Exemplary comparison of the open-loop and feedback solution. The opponent uses a sequential planning approach with $c_\mathrm{c}^2=10 c_\mathrm{c}^1$.}
	\label{fig:moving_horizon}
\end{figure}

If both players use the sequential planning approach, the mean overtaking times are \SI{8.2}{\second} and \SI{9.3}{\second}, respectively. The collision probability is approximately \SI{10}{\percent}. When the ego vehicle uses the interaction-aware approach, the collision probabilities decrease significantly for both the open-loop and the feedback solution. Furthermore, the ego vehicle can delay the overtaking maneuver of the opponent slightly for $\nicefrac{c_\mathrm{c}^2}{c_\mathrm{c}^1} = 1$, and significantly for $\nicefrac{c_\mathrm{c}^2}{c_\mathrm{c}^1} = 10$. The histograms in Fig.~\ref{fig:hist} visualize this observation. For the majority of lateral starting positions, the overtaking time remains at approximately \SI{8}{\second}; however for $\nicefrac{c_\mathrm{c}^2}{c_\mathrm{c}^1} = 10$, there is a significant share of scenarios where the game-theoretic planning approach can slow down the opponent, noticeably to a greater extent with the feedback solution. Fig.~\ref{fig:moving_horizon} shows an exemplary scenario where the open-loop and feedback results differ. With the open-loop solution in Fig.~\ref{fig:iQG_open_iLQR_open}, the opponent can successfully overtake the ego vehicle that only slightly swerves right, as to expect from the results in Section~\ref{sec:interaction_awareness_parameter}. With the feedback solution in Fig.~\ref{fig:iQG_closed_iLQR_open}, the ego vehicle can block the opponent, which decelerates to avoid a crash.

If the ego vehicle uses the sequential approach and the opponent uses the game theoretic approach, the overtaking times slightly decrease. The collision probabilities, however, remain unchanged. Only for $\nicefrac{c_\mathrm{c}^2}{c_\mathrm{c}^1} = 10$, the collision probability can be halved without the opponent having to sacrifice performance too much. This happens noticeably without a changed behavior of the ego vehicle compared to $\nicefrac{c_\mathrm{c}^2}{c_\mathrm{c}^1} = 1$ since $c_\mathrm{c}^2$ does not enter its cost function as in the example in Fig.~\ref{fig:both_ilqr} .

No collisions occur if both vehicles use the game theoretic approach and share the same solution concept. From both viewpoints, the other vehicle behaves as assumed according to the Nash equilibrium. The overtaking times are slightly smaller than the case where both use the sequential approach. Only for $\nicefrac{c_\mathrm{c}^2}{c_\mathrm{c}^1} = 10$, the ego vehicle can slow down the opponent for the feedback solution.

\section{Conclusion and Outlook}
\label{sec:dis_out}
The results of the previous section demonstrate the interaction awareness of the trajectory planning approach with \ac{iLQGame} in a head-to-head racing scenario on a straight race track. The formulation as a dynamic game couples the trajectories of the players through collision and progress costs, thereby combining prediction and planning. The resulting behavior is less yielding and even shows competitive blocking behavior for some configurations. This applies especially when the trailing vehicle follows a sequential approach. In this case, the frequency of collisions can also be reduced.

The cost distribution among the players plays a crucial role in influencing competitiveness and safety. We exemplified this by varying the collision costs of the trailing vehicle. As the collision cost increases, the leading vehicle can better slow down the trailing vehicle. Furthermore, the number of collisions decreases. Here, we assumed that both vehicles know about the increased collision cost of the trailing vehicle with the argument that the latter carries a greater responsibility to avoid collisions. Future work will analyze the cases where the vehicles have false assumptions about the others' cost function.

With the ability of \ac{iLQGame} to find open-loop and feedback solutions to the Nash equilibrium, we showed that the two strategy interpretations can result in fundamentally different behaviors. The leading vehicle tends to behave more competitively using the feedback than the open-loop solution, particularly when its collision costs are smaller. As a trailing vehicle, a tendency of the feedback solution outperforming the open-loop solution cannot be observed in our scenarios.

Summarizing, planning with \ac{iLQGame} exhibits interaction-aware behavior with greater competitiveness and safety than a sequential approach, especially in a defending position. Based on the simple scenarios analyzed here, a decision as to whether the open or the feedback solution is preferred cannot be made for general racing. However, we have shown that the two equilibrium concepts can lead to different behaviors. With our long-term goal to implement the game-theoretic planning approach in races involving full-scale prototypes, we will next analyze the performance on oval race tracks and road courses. Additionally, we will investigate outcomes in scenarios with $N\ge3$.


\bibliographystyle{IEEEtran}
\bibliography{IEEEabrv,literature}

\begin{thebibliography}{10}
\providecommand{\url}[1]{#1}
\csname url@rmstyle\endcsname
\providecommand{\newblock}{\relax}
\providecommand{\bibinfo}[2]{#2}
\providecommand\BIBentrySTDinterwordspacing{\spaceskip=0pt\relax}
\providecommand\BIBentryALTinterwordstretchfactor{4}
\providecommand\BIBentryALTinterwordspacing{\spaceskip=\fontdimen2\font plus
\BIBentryALTinterwordstretchfactor\fontdimen3\font minus
  \fontdimen4\font\relax}
\providecommand\BIBforeignlanguage[2]{{%
\expandafter\ifx\csname l@#1\endcsname\relax
\typeout{** WARNING: IEEEtran.bst: No hyphenation pattern has been}%
\typeout{** loaded for the language `#1'. Using the pattern for}%
\typeout{** the default language instead.}%
\else
\language=\csname l@#1\endcsname
\fi
#2}}

\bibitem{Rowold.2022}
M.~Rowold, L.~{\"O}gretmen, T.~Kerbl, and B.~Lohmann, ``{Efficient
  Spatiotemporal Graph Search for Local Trajectory Planning on Oval Race
  Tracks},'' \emph{{Actuators}}, vol.~11, no.~11, p. 319, 2022.

\bibitem{Raji.2022}
A.~Raji, A.~Liniger, A.~Giove, A.~Toschi, N.~Musiu, D.~Morra, M.~Verucchi,
  D.~Caporale, and M.~Bertogna, ``{Motion Planning and Control for Multi
  Vehicle Autonomous Racing at High Speeds},'' in \emph{{2022 IEEE 25th
  International Conference on Intelligent Transportation Systems
  (ITSC)}}.\hskip 1em plus 0.5em minus 0.4em\relax IEEE, 2022, pp. 2775--2782.

\bibitem{Jank.2023}
G.~Jank, M.~Rowold, and B.~Lohmann, ``{Hierarchical Time-Optimal Planning for
  Multi-Vehicle Racing},'' in \emph{{2023 International Conference on
  Intelligent Transportation Systems (ITSC)}}.\hskip 1em plus 0.5em minus
  0.4em\relax IEEE, 2023, accepted and presented.

\bibitem{Trautman2010}
P.~Trautman and A.~Krause, ``{Unfreezing the Robot: Navigation in Dense,
  Interacting Crowds},'' in \emph{{2010 IEEE/RSJ International Conference on
  Intelligent Robots and Systems}}.\hskip 1em plus 0.5em minus 0.4em\relax
  IEEE, 2010, pp. 797--803.

\bibitem{Schmidt2019}
M.~Schmidt, C.~Manna, J.~H. Braun, C.~Wissing, M.~Mohamed, and T.~Bertram,
  ``{An Interaction-Aware Lane Change Behavior Planner for Automated Vehicles
  on Highways Based on Polygon Clipping},'' \emph{{IEEE Robotics and Automation
  Letters}}, vol.~4, no.~2, pp. 1876--1883, 2019.

\bibitem{LeCleach2022}
S.~{Le Cleac'h}, M.~Schwager, and Z.~Manchester, ``{ALGAMES: a fast augmented
  Lagrangian solver for constrained dynamic games},'' \emph{{Autonomous
  Robots}}, vol.~46, no.~1, pp. 201--215, 2022.

\bibitem{FridovichKeil2020}
D.~Fridovich-Keil, E.~Ratner, L.~Peters, A.~D. Dragan, and C.~J. Tomlin,
  ``{Efficient Iterative Linear-Quadratic Approximations for Nonlinear
  Multi-Player General-Sum Differential Games},'' in \emph{{2020 IEEE
  International Conference on Robotics and Automation (ICRA)}}.\hskip 1em plus
  0.5em minus 0.4em\relax IEEE, 2020, pp. 1475--1481.

\bibitem{Crosato2023}
L.~Crosato, H.~P.~H. Shum, E.~S.~L. Ho, and C.~Wei, ``{Interaction-Aware
  Decision-Making for Automated Vehicles Using Social Value Orientation},''
  \emph{{IEEE Transactions on Intelligent Vehicles}}, vol.~8, no.~2, pp.
  1339--1349, 2023.

\bibitem{Fisac2019}
J.~F. Fisac, E.~Bronstein, E.~Stefansson, D.~Sadigh, S.~S. Sastry, and A.~D.
  Dragan, ``{Hierarchical Game-Theoretic Planning for Autonomous Vehicles},''
  in \emph{{2019 International Conference on Robotics and Automation
  (ICRA)}}.\hskip 1em plus 0.5em minus 0.4em\relax IEEE, 2019, pp. 9590--9596.

\bibitem{Bhargav2021}
\BIBentryALTinterwordspacing
J.~Bhargav, J.~Betz, H.~Zheng, and R.~Mangharam, ``{Track based Offline Policy
  Learning for Overtaking Maneuvers with Autonomous Racecars}.'' [Online].
  Available: \url{10.48550/arXiv.2107.09782}
\BIBentrySTDinterwordspacing

\bibitem{Zheng2022}
\BIBentryALTinterwordspacing
H.~Zheng, Z.~Zhuang, J.~Betz, and R.~Mangharam, ``{Game-theoretic Objective
  Space Planning}.'' [Online]. Available: \url{10.48550/arXiv.2209.07758}
\BIBentrySTDinterwordspacing

\bibitem{Liniger2020}
A.~Liniger and J.~Lygeros, ``{A Noncooperative Game Approach to Autonomous
  Racing},'' \emph{{IEEE Transactions on Control Systems Technology}}, vol.~28,
  no.~3, pp. 884--897, 2020.

\bibitem{Wang2021}
M.~Wang, Z.~Wang, J.~Talbot, J.~C. Gerdes, and M.~Schwager, ``{Game-Theoretic
  Planning for Self-Driving Cars in Multivehicle Competitive Scenarios},''
  \emph{{IEEE Transactions on Robotics}}, vol.~37, no.~4, pp. 1313--1325, 2021.

\bibitem{Wang.2019}
Z.~Wang, R.~Spica, and M.~Schwager, ``{Game Theoretic Motion Planning for
  Multi-robot Racing},'' in \emph{{Distributed Autonomous Robotic Systems}},
  ser. {Springer Proceedings in Advanced Robotics}, N.~Correll, M.~Schwager,
  and M.~Otte, Eds.\hskip 1em plus 0.5em minus 0.4em\relax Cham: {Springer
  International Publishing}, 2019, vol.~9, pp. 225--238.

\bibitem{Spica.2020}
R.~Spica, E.~Cristofalo, Z.~Wang, E.~Montijano, and M.~Schwager, ``{A Real-Time
  Game Theoretic Planner for Autonomous Two-Player Drone Racing},'' \emph{{IEEE
  Transactions on Robotics}}, vol.~36, no.~5, pp. 1389--1403, 2020.

\bibitem{MAYNE1966}
D.~Q. Mayne, ``{A Second-order Gradient Method for Determining Optimal
  Trajectories of Non-linear Discrete-time Systems},'' \emph{{International
  Journal of Control}}, vol.~3, no.~1, pp. 85--95, 1966.

\bibitem{Li2004}
W.~Li and E.~Todorov, ``{Iterative Linear Quadratic Regulator Design for
  Nonlinear Biological Movement Systems},'' in \emph{{Proceedings of the First
  International Conference on Informatics in Control, Automation and
  Robotics}}.\hskip 1em plus 0.5em minus 0.4em\relax {SciTePress - Science and
  and Technology Publications}, 2004, pp. 222--229.

\bibitem{Basar1999}
T.~Ba{\c{s}}ar and G.~J. Olsder, \emph{{Dynamic Noncooperative Game Theory}},
  2nd~ed., ser. {Classics in applied mathematics}.\hskip 1em plus 0.5em minus
  0.4em\relax Philadelphia, Pa.: {SIAM Soc. for Industrial and Applied
  Mathematics}, 1999, vol.~23.

\bibitem{Schwarting2021}
W.~Schwarting, A.~Pierson, S.~Karaman, and D.~Rus, ``{Stochastic Dynamic Games
  in Belief Space},'' \emph{{IEEE Transactions on Robotics}}, vol.~37, no.~6,
  pp. 2157--2172, 2021.

\bibitem{Kavuncu.2021}
T.~Kavuncu, A.~Yaraneri, and N.~Mehr, ``{Potential iLQR: A Potential-Minimizing
  Controller for Planning Multi-Agent Interactive Trajectories},'' in
  \emph{{Robotics: Science and Systems XVII}}.\hskip 1em plus 0.5em minus
  0.4em\relax {Robotics: Science and Systems Foundation}, 2021.

\bibitem{Zhu.2023}
E.~L. Zhu and F.~Borrelli, ``{A Sequential Quadratic Programming Approach to
  the Solution of Open-Loop Generalized Nash Equilibria},'' in \emph{{2023 IEEE
  International Conference on Robotics and Automation (ICRA)}}.\hskip 1em plus
  0.5em minus 0.4em\relax IEEE, 2023, pp. 3211--3217.

\bibitem{Starr1969b}
A.~W. Starr and Y.~C. Ho, ``{Further properties of nonzero-sum differential
  games},'' \emph{{Journal of Optimization Theory and Applications}}, vol.~3,
  no.~4, pp. 207--219, 1969.

\bibitem{Rowold.2023}
M.~Rowold, L.~{\"O}gretmen, U.~Kasolowsky, and B.~Lohmann, ``{Online
  Time-Optimal Trajectory Planning on Three-Dimensional Race Tracks},'' in
  \emph{{2023 IEEE Intelligent Vehicles Symposium (IV)}}.\hskip 1em plus 0.5em
  minus 0.4em\relax IEEE, 2023, pp. 1--8.

\bibitem{Chen2017}
J.~Chen, W.~Zhan, and M.~Tomizuka, ``{Constrained iterative LQR for on-road
  autonomous driving motion planning},'' in \emph{{2017 IEEE 20th International
  Conference on Intelligent Transportation Systems (ITSC)}}.\hskip 1em plus
  0.5em minus 0.4em\relax IEEE, 2017, pp. 1--7.

\end{thebibliography}

\section*{APPENDIX}
\label{sec:appendix}
The derivations of the recursions without linear cost terms are given in \cite{Basar1999}. The supplementary material to \cite{FridovichKeil2020} derives the following recursions \eqref{eq:feedback_recursion} and \eqref{eq:open_loop_recursion} with linear cost terms\footnote{\href{https://github.com/HJReachability/ilqgames/tree/master/derivations}{https://github.com/HJReachability/ilqgames/tree/master/derivations}}.

\subsection{Recursion for the Feedback Solution}
\label{sec:recursion_feedback}
Beginning with $\bm{P}_K^i = \bm{Q}_K^i$ and $\bm{p}_K^i = \bm{q}_K^i$, calculate:\\
For $k$ from $K-1$ to $0$:
\begin{subequations}
	\label{eq:feedback_recursion}\\
	\begin{alignat}{2}
		& \bm{F}_k = \bm{A}_k - \sum_{j=1}^{N}\bm{B}_k^j\bm{K}_k^j \text{, \quad} \bm{\beta}_k = - \sum_{j=1}^{N}\bm{B}_k^j\bm{k}_k^j\\
		& \bm{P}_k^i = \bm{Q}_k^i + \bm{F}_k^\top\bm{P}_{k+1}^i\bm{F}_k + \sum_{j=1}^{N}\bm{K}_k^{j\top}\bm{R}_k^{ij}\bm{K}_k^j\label{eq:coupled_riccati}\\
		&\begin{aligned}
			\bm{p}_k^i = \bm{q}_k^i + &\bm{F}_k^\top\left(\bm{p_{k+1}^i} + \bm{P}_{k+1}^i\bm{\beta}_k\right) + \\ &\sum_{j=1}^{N}\bm{K}_k^{j\top}\bm{R}_k^{ij}\bm{k}_k^j - \bm{K}_k^{j\top}\bm{r}_k^{ij}\text{.}
		\end{aligned}
	\end{alignat}
\end{subequations}
\subsection{Recursion for the Open-Loop Solution}
\label{sec:recursion_open_loop}
Beginning with $\bm{M}_K^i = \bm{Q}_K^i$ and $\bm{m}_K^i = \bm{q}_K^i$, calculate:\\
For $k$ from $K-1$ to $0$:
\begin{subequations}
	\label{eq:open_loop_recursion}\\
	\begin{alignat}{2}
		&\bm{\Lambda}_k = \mathbb{I} + \sum_{j=1}^{N}\bm{B}_k^j\bm{R}_k^{jj^{-1}}\bm{B}_k^{j\top}\bm{M}_{k+1}^j\\
		&\begin{aligned}
			\bm{m}_k^i = &\bm{A}_k^\top\Biggl[\bm{m}_{k+1}^i-\bm{M}_{k+1}^i\bm{\Lambda}_k^{-1}\cdot\Biggr.\\
			\Biggl.&\sum_{j=1}^{N}\bm{B}_k^j\bm{R}_k^{jj^{-1}}\left(\bm{B}_k^{j\top}\bm{m}_{k+1}^j+\bm{r}_k^{jj}\right)\Biggr]+\bm{q}_k^i
		\end{aligned}\\
		&\bm{M}_k^i = \bm{Q}_k^i+\bm{A}_k^\top\bm{M}_{k+1}^i\bm{\Lambda}_k^{-1}\bm{A}_k\text{.}
	\end{alignat}
\end{subequations}

\end{document}